\documentclass{article}

\usepackage{microtype}
\usepackage{graphicx}
\usepackage{subfigure}
\usepackage{booktabs} 

\usepackage{hyperref}

\usepackage[accepted]{icml2024}

\usepackage{amsmath}
\usepackage{amssymb}
\usepackage{mathtools}
\usepackage{amsthm}
\usepackage{tabularx}

\usepackage[capitalize,noabbrev]{cleveref}

\theoremstyle{plain}

\theoremstyle{definition}

\theoremstyle{remark}

\usepackage[textsize=tiny]{todonotes}

\icmltitlerunning{DFORM: Diffeomorphic vector field alignment for assessing dynamics across learned models}

\begin{document}

\twocolumn[

\icmltitle{DFORM: Diffeomorphic vector field alignment for \\ assessing dynamics across learned models}

\icmlsetsymbol{equal}{*}

\begin{icmlauthorlist}
\icmlauthor{Ruiqi Chen}{yyy}
\icmlauthor{Giacomo Vedovati}{yyy}
\icmlauthor{Todd Braver}{xxx}
\icmlauthor{ShiNung Ching}{yyy}
\end{icmlauthorlist}

\icmlaffiliation{yyy}{Department of Electrical and Systems Engineering, Washington University in St. Louis, Missouri, USA}
\icmlaffiliation{xxx}{Department of Psychological and Brain Sciences, Washington University in St. Louis, Missouri, USA}

\icmlcorrespondingauthor{ShiNung Ching}{shinung@wustl.edu}

\icmlkeywords{Dynamical systems, Vector field alignment, Diffeomorphism, Residual networks}

\vskip 0.3in
]


\printAffiliationsAndNotice{}  

\begin{abstract}
Dynamical system models such as Recurrent Neural Networks (RNNs) have become increasingly popular as hypothesis-generating tools in scientific research. Evaluating the dynamics in such networks is key to understanding their learned generative mechanisms.
However, comparison of learned dynamics across models is challenging due to their inherent nonlinearity and because \textit{a priori} there is no enforced equivalence of their coordinate systems.  Here, we propose the DFORM (Diffeomorphic vector field alignment for comparing dynamics across learned models) framework. DFORM learns a nonlinear coordinate transformation which provides a continuous, maximally one-to-one mapping between the trajectories of learned models, thus approximating a diffeomorphism between them. 
The mismatch between DFORM-transformed vector fields defines the orbital similarity between two models, thus providing a generalization of the concepts of smooth orbital and topological equivalence. As an example, we apply DFORM to models trained on a canonical neuroscience task, showing that learned dynamics may be functionally similar, despite overt differences in attractor landscapes.
\end{abstract}

\section{Introduction}

Recent progress in theoretical neuroscience has seen significant advances in the generation of recurrent neural networks as surrogates for brain networks (e.g., \cite{yang2019task, sussillo2009generating, barak2017recurrent}), in an effort to generate hypotheses regarding how the brain implements various functions \cite{sussillo2014neural}. An underlying question in the analysis of said networks, then, is to elucidate the mechanisms -- and specifically the dynamics -- that they embed \cite{maheswaranathan2019universality}. In this regard, it is insufficient to simply compare the outputs of these networks, since such outputs are often low dimensional projections that are opaque with respect to underlying generative mechanisms. Instead, a full characterization of learned dynamics must be based on the vector field that defines such dynamics. However, these vector fields emerge via stochastic learning processes that \textit{a priori} do not maintain fixed coordinate systems and, moreover, are able to introduce arbitrary nonlinear transformations between state variables. 
The net result is that vector fields may be distorted, permuted or combined in rather arbitrary ways across models, despite implementing similar dynamics and functions.
Given such fundamental difficulties, how can we compare the dynamics of different models, in order to better understand commonalities between them?

\begin{figure}[tb]
\vskip 0.2in
\begin{center}
    \centerline{\includegraphics[width=0.47\textwidth]{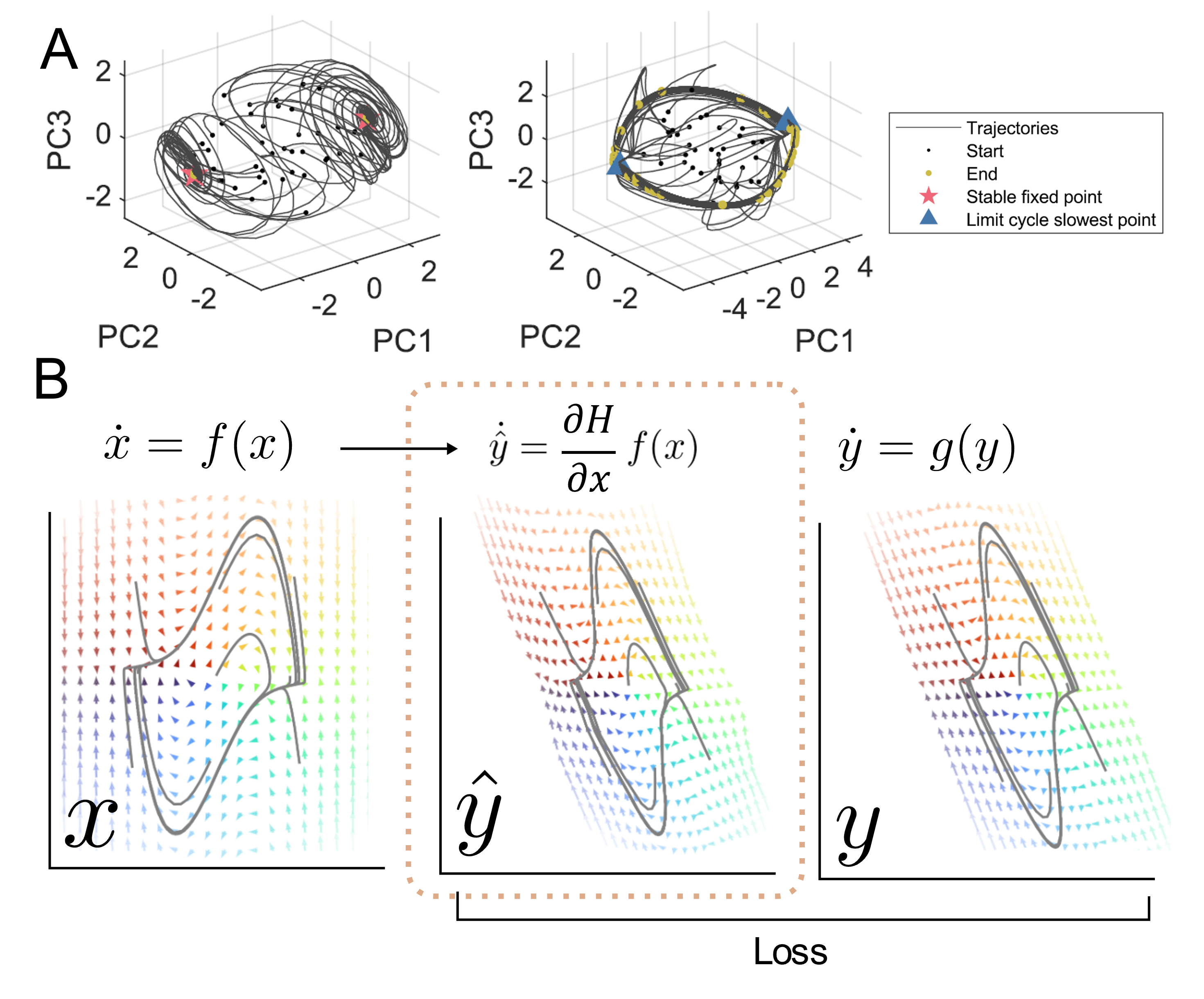}}
    \caption{DFORM Schematic. \textbf{A.} Many efforts to compare leaned models resort to dimensionality reduction and assessment of limit sets (e.g., attractors), often by visualization. \textbf{B.} We propose DFORM to learn a diffeomorphism that directly transforms vector fields, thus allowing for rigorous assessment of their similarity.}
    \label{fig:schematic}
\end{center}
\vskip -0.2in
\end{figure}

The most straightforward approach to this issue has been to simply perform dimensionality reduction, e.g., via principal components analysis, on learned models and then visualize their ensuing trajectories or orbits (\cref{fig:schematic}A). While such an approach can illuminate the limit sets and hence attractor landscape of such models, it carries several important shortcomings. First, it predominantly characterizes the asymptotic behavior of models (e.g., stable fixed points, limit cycles), and is less sensitive to unstable structures such as saddle points. Further, it relies on numerically solving the learned models from sampled initial conditions, in order to generate a set of orbits from which to build the attractor landscape in question. Finally, the approach lacks a clear quantification for the correspondence between landscapes or models, and often reduces to visual inspection or extraction of specific features (e.g., number of fixed points).

The goal of this paper is to introduce an approach to directly compare the dynamics of learned models via the geometry of their vector fields, without necessarily relying on simulation and dimensionality reduction. In this regard, we appeal to the fundamental, dynamical systems theoretic notion of smooth orbital equivalence. 
Two systems are smoothly orbitally equivalent if there exists a diffeomorphism between the two phase space that matches the orbits (trajectories) of one system and to those of the other in a one-to-one fashion. 
In other words, they have qualitatively the same dynamics. Note that systems that are orbitally equivalent are also topologically equivalent and share the same number of and type of limit sets (fixed points, limit cycles).
However, for all but the simplest of systems, finding a diffeomorphism to verify smooth orbital equivalence is highly nontrivial. Even if this were possible, equivalence is all or none rather than a continuous measure of functional similarity. For instance, two systems on either side of a bifurcation may have geometrically quite similar vector fields, but would not be equivalent in a strict sense.
As a result, this notion, while fundamentally rigorous and sound, has not been operationalized in practice.

Here, we directly address the problem of smooth orbital equivalence by learning an orbit-matching coordinate transformation between two systems through invertible residual networks \cite{behrmannInvertibleResidualNetworks2019}. To this end, we derive a first-principles loss function using the Lie derivative of the transformation along the orbits of the systems under comparison (\cref{fig:schematic}B). Such loss can be back-propagated all the way through the ordinary differential equations (ODEs) governing the dynamic systems, and it can be minimized end-to-end without any numerical integration. Further, we propose a bidirectional architecture and training scheme that facilitates generalization while also providing an analytic, differentiable approximation of the inverse transformation. This, combined with analytical properties of residual networks makes the learned function almost diffeomorphic. Crucially, the end result generalizes the notion of orbital equivalence into a continuous similarity metric that characterizes how well one system's vector field geometrically maps to that of the other. We demonstrate the efficacy of our approach on canonical examples of systems with known equivalence, and furthermore illustrate the ability of the technique to discover functional consistency in learned large-scale recurrent (dynamical) neural network models.

\section{Background and related work}

\subsection{Comparison of dynamical systems}

As mentioned above, a common way to characterize and compare high-dimensional dynamical systems is to numerically integrate the trajectories (i.e., by means of an ODE solver) and assess the limit sets. While this approach can surface asymptotically attractive sets in state space, identifying all equilibria including the saddle nodes remains challenging \cite{golub2018fixedpointfinder, katzUsingDirectionalFibers2018}. Similarly, models can be compared based on the number of type of limit sets they embed, usually with the help of dimensionality reduction analysis and visualization. The singular vectors canonical correlations analysis (SVCCA) framework \cite{raghu2017svcca}, for example, finds a one-dimensional projection of the simulated trajectories for each system, such that the correlation between two systems' projections are maximized.
However, such comparison is either qualitative (in the case of visualization) or limited insofar as it can only asses `strict' topological equivalency. In \cite{smith2021reverse}, a topological similarity measure is proposed by computing a transition probability matrix associated with traversal of stable fixed points, where the transition probability is obtained by perturbing around these fixed points. 

Going beyond limit sets, the MARBLE framework \cite{gosztolai2023interpretable} proposes a similarity measure based on local vector field features (vectors and higher-order derivatives) sampled from across the phase space. The features are embedded into lower-dimensional space through contrastive learning, and the embedded distributions from different systems can be compared using an optimal transport distance. MARBLE is also model-free but can be adapted to use the model-based features as well.

To the best of our knowledge, the method most directly related to the current paper is the recent Dynamical Similarity Analysis (DSA) framework \cite{ostrow2023beyond}. DSA learns a linear coordinate transformation which maximizes the cosine similarity between the vector fields of two linear time-invariant systems. For nonlinear dynamical systems, such linear systems can be obtained through Dynamical Mode Decomposition (DMD), which approximates its Koopman eigenspectrum \cite{schmid2022dynamic}. Importantly, topologically conjugate systems would have the same Koopman eigenspectrum. Therefore, assuming that the DMD-approximated linear systems well-characterizes the Koopman eigenspectrum, DSA quantifies how far the two systems are from being conjugate. Despite shared motivation aspects, our work has important distinctions from DSA along conceptual and methodological dimensions. We work directly in the state space of the original nonlinear systems, i.e., without any linearization. This means we do not garner any approximation error from DMD or similar basis expansions.
Second, we provide an analytic approximation of the actual diffeomorphism between the two systems, while in DSA such a diffeomorphism can only be computed when the DMD mapping can be inverted, which is generally challenging \cite{bruntonKoopmanInvariantSubspaces2016}. Indeed, direct matching of two topologically equivalent dynamical systems through Koopman eigenfunctions is a difficult problem due to the geometric multiplicity of Koopman eigenvalues \cite{bolltMatchingEvenRectifying2018}.

\subsection{Residual networks as diffeomorphisms}

Constructing a diffeomorphism to minimize the discrepancy between two \textit{scalar} fields has been studied extensively, particularly in image registration problems \cite{begComputingLargeDeformation2005}. The  development of residual networks (ResNet) \cite{heDeepResidualLearning2015} offers a powerful functional approximation tool for such learning problems, and we will exploit this architecture in our treatment of vector fields. It is known that the layers of a ResNet can be considered as Euler-discretization of the integration of a flow of a diffeomorphism \cite{rousseauResidualNetworksFlows2020}, or so-called Neural ODE \cite{chenNeuralOrdinaryDifferential2018, marionImplicitRegularizationDeep2023}. The invertibility of the ResNet mapping can be enforced as a regularization term \cite{rousseauResidualNetworksFlows2020}, or more straightforwardly, by constraining the Lipschitz constant of the residual blocks \cite{behrmannInvertibleResidualNetworks2019, goukRegularisationNeuralNetworks2021}.

ResNet-based registration has been successfully applied to one-dimensional (time warping) \cite{huangResidualNetworksFlows2021}, two-dimensional (images) \cite{amorResNetLDDMMAdvancingLDDMM2023}, and high-dimensional (point clouds) \cite{battikhKNNResResidualNeural2023} scalar fields. However, to our knowledge, it has never been used on vector fields. Importantly, as will be shown below, while the mismatch loss function for scalar fields only involves the output of the ResNet, \textit{a proper mismatch loss function for vector fields contains the Lie derivative of the ResNet with respect to the fields}. Whether such loss function can be effectively minimized end-to-end has never been demonstrated before and constitutes a key innovation in the current work.

\section{Theory}
We consider the problem is evaluating the smooth orbital equivalency of learned dynamical models, such as recurrent neural networks. We denote any two such networks as
\begin{equation}
    \dot x = f(x)
\end{equation}
and
\begin{equation}
    \dot y = g(y),
\end{equation}
where $x,y \in\mathbb{R}^N$.
From a dynamical systems perspective, these networks are said to be smoothly orbitally equivalent, should there exist a diffeomorphism $\mathcal{H}: \mathbb{R}^N \to \mathbb{R}^N$ such that 
\begin{equation}
    \frac{\partial \mathcal{H}}{\partial x}f(x) = c(x)g(\mathcal{H}(x))
\end{equation}
where $c(x) > 0$ is a smooth function. In practice, the evaluation of $\mathcal{H}$ is intractable for all but the simplest of dynamical systems.

In order to operationalize this comparison, we propose DFORM, in which a feedforward network is trained to serve as $\mathcal{H}$.  Denoting this network as the transformation $H: \mathbb{R}^N \to \mathbb{R}^N$, we define
\begin{equation}
    \hat y = H(x)
\end{equation}
Our goal is to learn $H$ such that the dynamics of $\hat y$ recapitulate those of $y$. For this we introduce the \textit{orbital similarity loss}, based on the Lie derivative of $H$ with respect to the flow $f(\cdot)$:
\begin{equation}\label{topoloss}
    J^{l}_{f, g, H}(x) = \left\| \frac{\frac{\partial H}{\partial x}f(x)}{\left\|\frac{\partial H}{\partial x}f(x)\right\|_{2}} - \frac{g(H(x))}{\|g(H(x))\|_{2}} \right\|_2^2
\end{equation}
This loss captures the geometric mis-alignment of the vector field $f(x)$ and the vector field for $y$ subject to $H$.
Specifically, define the \textit{push-foward} of $f$ by $\mathcal{H}$ as $\mathcal{H}_{*}f$ such that
\begin{equation}
    \mathcal{H}_{*}f(\hat y) := \frac{\partial H}{\partial x}f(H^{-1}(\hat y))
\end{equation}
The loss function can be rewritten simply as
\begin{equation}
    J^{l}_{f, g, H}(H^{-1}(\hat{y})) = \left\| \frac{\mathcal{H}_{*}f(\hat y)}{\left\|\mathcal{H}_{*}f(\hat y)\right\|_2} - \frac{g(\hat{y})}{\|g(\hat{y})\|_{2}} \right\|_2^2
\end{equation}
which is the squared normalized Euclidean distance between the push-forward and target vector fields evaluated at $\hat{y} = Hx$.
The loss is minimized when $\mathcal{H}_{*}f(\hat y) = c(\hat y)g(\hat y)$ where $c(\hat y) > 0$, i.e., when the push-forward and target vector fields differ only in magnitude but not direction. In this case, $H$ exactly maps the $x$ orbits to the $y$ orbits in a one-to-one fashion.

\subsection{Orbital similarity} \label{sec:orbit}
DFORM provides a generalization of the notion of orbital equivalence. 
Specifically, after applying DFORM across two systems $\dot{x} = f(x)$ and $\dot{y} = g(y)$, we can define the similarity between the two vector fields $f$ and $g$ as:
\begin{equation}
    \textbf{E}_{x}\left[\text{cos}\angle\left(\frac{\partial H}{\partial x}f(x), g(H(x))\right)\right]
\end{equation}
where $\angle(\mathbf{v}, \mathbf{w})$ indicates the angle between two vectors $\mathbf{v}$ and $\mathbf{w}$. The outcome of this evaluation will heavily depend on the distribution of $x$. In practice, we found that when $f$ is similar to $g$ over a constrained domain (e.g., when $f$ has a single equilibrium and $g$ has three), DFORM will sometimes squeeze all orbits in $f$ into said domain, generating high similarity if we sample $x$ from the domain of $f$. Therefore, we also repeat the calculation using $H^{-1}(y)$ where $y$ is sampled from the domain of $g$. We retain the lower value in these two (sampling from the domain of $f$ or $g$) as the final result.

\section{Implementation}

\subsection{Model architecture}

There are two key ideas that enable us to solve the problem at hand. First, we leverage a residual network architecture to serve as $H$. A ResNet contains multiple layers that can be written as:
\begin{equation}
    x_{l + 1} = f_{l}(x_{l}) = x_{l} + g_{l}(x_{l})
\end{equation}
where $l$ is the layer index. In \cite{behrmannInvertibleResidualNetworks2019}, it is proved that $f_{l}$ is bijective if $g_{l}$ has a Lipschitz constant smaller than one. The inverse can be numerically determined through an iteration procedure with guaranteed exponential convergence. Moreover, the inverse function is also provably Lipschitz continuous, thus differentiable almost everywhere, according to Rademacher's theorem. Therefore, while not strictly diffeomorphic, the mapping learned by such an invertible ResNet (i-ResNet) is still bijective, continuous, and has bounded rate-of-change in both directions. In our experiments, unless stated otherwise, we used a ResNet with 10 layers, where each residual block $g_{l}$ contains a fully connected sub-layer, a ReLU activation sub-layer and another fully connected sub-layer. The three sub-layers are of width $(n, 2n, n)$ respectively, where $n$ is the dimension of the to-be-compared systems. The Lipschitz condition was satisfied by scaling the spectral norm of connection weights to below $0.99$ after each weight update \cite{goukRegularisationNeuralNetworks2021}.

Second, we require a mechanism to effectively minimize the orbital similarity loss \eqref{topoloss} across the domain of \textit{both} systems, such that the regions-of-interest (ROIs) in both systems can be well-described. When training a single i-ResNet to map $y = H(x)$, the orbital similarity loss is evaluated only at points $y$ whose inverse $x = H^{-1}(y)$ has been sampled, which often constitute only a small portion of the ROI in $y$'s domain due to the problem mentioned in \cref{sec:orbit}. Consequently, $H^{-1}$ generalizes poorly outside the range of sampled $H(x)$.

To address this issue, we propose a bidirectional modeling architecture which exploits the symmetry in registration problems. Instead of training one mapping, we train two i-ResNets $\phi(x) = y$ and $\psi(y) = x$ simultaneously while regularizing them to be the inverse of each other. Define the two systems to be matched as $f, g$ and their sample distribution as $p(x), q(y)$. During each training batch, $\phi$ and $\psi$ are updated in a symmetrical way. We draw $S$ ($S = 128$ throughout the current paper) samples from the domain of the first function, e.g., $\phi$, and calculate a "forward" orbital similarity loss $J_{f} = J^{l}_{f, g, \phi}(x)$. Treating $\phi(x)$ as samples from the second domain, we calculate another "backward" orbital similarity loss $J_{b} = J^{l}_{g, f, \psi}(\phi(x))$. Minimizing this term ensures that $\psi$ (i.e., the approximation of $\phi^{-1}$) generalizes well to the images of $x$, which might be quite different from samples from $q$. A third term regularizes the two networks to be the inverse of each other: $J_{i} = \|x - \psi(\phi(x))\|^{2}_{2}$. The total loss during this update is thus:
\begin{equation}
    J_{\phi} = w_{f}J_{f} + w_{b}J_{b} + w_{i}J_{i}
\end{equation}
where the regularization weights $w_{f}, w_{b}, w_{i}$ were all set to one. In practice, we found that removing the backward or inverse loss would both impact the performance of DFORM. The procedure described above is summarized in \cref{alg:uni}. During each training batch, the procedure is repeated in symmetry for $\psi$, as described in \cref{alg:training}.

\begin{algorithm}[tb]
    \caption{One training batch for DFORM}
    \label{alg:uni}
\begin{algorithmic}
    \STATE {\bfseries Input:} systems $f(x), g(y)$, sample distributions $p(x), q(y)$, i-ResNets $\phi: \phi(x) = y, \psi: \psi(y) = x$
    \STATE {\bfseries Output:} Updated $\phi, \psi$
    \STATE {\bfseries Hyperparameters:} regularization weights $w_{f}, w_{b}, w_{i}$, batch size $N$, Lipschitz constant $c$
    \STATE Draw $N$ samples $x \sim p(x)$
    \STATE Calculate $\hat{y} = \phi(x)$
    \STATE Calculate forward loss $J_{f} = J^{l}_{f, g, \phi}(x)$
    \STATE Calculate backward loss $J_{b} = J^{l}_{g, f, \psi}(\hat{y})$
    \STATE Calculate inverse loss $J_{i} = \text{MSE}(x, \psi(\phi(x)))$
    \STATE Calculate total loss $J = w_{f}J_{f} + w_{b}J_{b} + w_{i}J_{i}$
    \STATE Update weights by back-propagating $J$
    \STATE Scale weights to have spectral norm smaller than $c$
\end{algorithmic}
\end{algorithm}

\begin{algorithm}[tb]
    \caption{DFORM Training}
    \label{alg:training}
\begin{algorithmic}
    \STATE {\bfseries Input:} systems $f(x), g(y)$, sample distributions $p(x), q(y)$, training time $m$
    \STATE {\bfseries Output:} i-ResNets $\phi: \phi(x) = y, \psi: \psi(y) = x$
    \STATE Initialize $\phi, \psi$
    \FOR{$i=1$ {\bfseries to} $m$}
        \STATE Train($f$, $g$, $p$, $q$, $\phi$, $\psi$) through \cref{alg:uni}
        \STATE Train($g$, $f$, $q$, $p$, $\psi$, $\phi$) through \cref{alg:uni}
    \ENDFOR
\end{algorithmic}
\end{algorithm}

Although analytically complicated, the gradient of such loss function with respect to the parameters of $H$ can be obtained easily through automatic differentiation, as implemented in \textit{PyTorch}. We used the Adam \cite{kingmaAdamMethodStochastic2017} optimizer with a learning rate of $0.001$ to minimize the loss. The models were trained on a single V100 GPU for several hundred to several thousand batches depending on the dimension of the system, and the run time is typically several minutes.

\subsection{Selection of sample distributions}

A key component in measuring orbital similarity is to determine a proper sample distribution, which influences not only the similarity metric but also its interpretation. For a bounded phase space, a uniform distribution is a reasonable choice. However, more commonly, the phase space of the system under investigation is unbounded, and we inevitably need to weight some area in the phase space more than others. We argue that the choice of sample distribution depends on the question of interest: for analytically derived systems where the similarity is evaluated `in a vacuum', a normal or uniform distribution over a large area (particularly covering the limit sets) might be suitable. However, for dynamical models derived from empirical data or trained on tasks, e.g., computational models of the brain, the similarity calculation should also follow the distribution of the data (either observed or simulated). Such choice ensures that the estimated similarity is \textit{functionally relevant}, leading to better interpretations about common computational mechanisms across different models. In particular, for dynamical models of biological systems such as the brain, the asymptotic distribution of hidden states under the influence of noise reflects a `working dynamical regime'. In this sense, the DFORM orbital similarity becomes a noisy measure of topological equivalence, where the discrepancy between push-forward and target vector field might be associated with the errors inherent to the observation or modeling processes.

\section{Results}

\subsection{DFORM can identify the coordinate transformation between topologically equivalent systems}

\subsubsection{Nonlinear systems under linear transformation}

First, we demonstrate the ability of DFORM to establish topological equivalence between two systems that are associated with each other by a linear coordinate transformation. Here we adopted two classical nonlinear systems in $\mathbb{R}^2$ with nontrivial limit sets. The first one is a Van der Pol oscillator:
\begin{equation}\label{VDP}
    \begin{cases}
        \dot{x_1} = x_2 \\
        \dot{x_2} = \mu (1 - x^{2})x_2 - x_1
    \end{cases}
\end{equation}
which has one non-harmonic stable limit cycle when $\mu > 0$. Here we sampled $\mu$ uniformly between $1.5$ and $3.5$. The other system is the normal form of a supercritical pitchfork bifurcation:
\begin{equation}\label{Pitch}
    \begin{cases}
        \dot{x_1} = \mu x_1 + x_1^3 \\
        \dot{x_2} = -x_2
    \end{cases}
\end{equation}
which has one saddle node at the origin and two stable equilibria at $(\pm \sqrt{\mu}, 0)^T$ when $\mu > 0$. Here, we sampled $\mu$ uniformly between $2$ and $4$.

We sampled 30 systems of each type. For each system $f_1$, we generated a random matrix $Q \in \mathbb{R}^{2\times 2}$ whose entries were sampled from standard normal distribution. We ensured the determinant of $Q$ to be positive, although this requirement is likely unnecessary. A topologically equivalent system $f_2$ was defined by $f_{2}(\mathbf{y}) := Qf_{1}(Q^{-1}\mathbf{y})$, and the two systems are related by $\mathbf{y} = Q\mathbf{x}$.

We used DFORM to identify the equivalence between $f_1$ and $f_2$ for each system realization. The DFORM ResNet was trained for 2000 batches, and the highest similarity among three repetitions was retained. To facilitate learning the limit set structure of the systems, samples from the Van der Pol systems were drawn from a uniform distribution $x_1 \in [-3, 3]$, $x_2 \in [-4, 4]$ which covers the limit cycle for most systems; samples from the pitchfork systems were drawn from uniform distribution $x_1 \in [-1.5 \sqrt{\mu}, 1.5 \sqrt{\mu}]$ and $x_2 \in [-1, 1]$, which covers all fixed points.

Results are shown in \cref{tab:nonlinear-sys}. In 22 of 30 Van der Pol systems and 26 of 30 pitchfork systems, DFORM was able to obtain a similarity higher than $0.99$. A specific example is shown in \cref{fig:VDP}.

\subsubsection{Large RNNs under linear transformation}

To examine how well DFORM scales to higher-dimensional settings, we applied DFORM to 64-dimensional `vanilla' RNNs, which have been widely used in theoretical neuroscience \cite{yildizRevisitingEchoState2012, maassRealtimeComputingStable2002}. Dynamics of these models are given by:
\begin{equation}
    \dot{x} = -x + W\text{tanh}(x)
\end{equation}
where $x\in \mathbb{R}^n$ is the neuronal hidden state ($n = 64$) and $W\in\mathbb{R}^{n\times n}$ is the connectivity matrix. Following \cite{mastrogiuseppeLinkingConnectivityDynamics2018}, we enforced a `random plus low rank' structure on the connectivity matrix: $W = J + mn^\top$. Such system can manifest nontrivial attractors induced by the low rank structure \cite{schuesslerDynamicsRandomRecurrent2020}. Here $J \in \mathbb{R}^{n\times n}$ is sampled from a normal distribution with standard deviation 0.5. $m\in \mathbb{R}^{n\times k}$ ($k = 2$) is sampled from a standard normal distribution. $n\in\mathbb{R}^{n\times k}$ is sampled from normal distribution with standard deviation $1/\sqrt{n}$. We generated 30 random RNNs. For each, we applied a random orthogonal coordinate transformation to generate a topologically similar model. DFORM was trained between original and transformed systems for 6000 batches, sampling from the standard normal distribution. Across 30 instances, DFORM was able to generate a mean similarity above $0.9$, with a minimum of $0.890$.

\begin{table}[htb]

\vskip 0.15in
\begin{center}
\begin{small}
\begin{sc}
\begin{tabularx}{.99\columnwidth}{lcccr}
\toprule
Type & Mean & Median \\
\midrule  
VDP oscillator & $0.915 \pm 0.026$ & $0.997 \pm 0.146$ \\
Bistable pitchfork & $0.919 \pm 0.038$ & $0.998 \pm 0.002$ \\
64D RNN & $0.925 \pm 0.006$ & $0.927 \pm 0.057$ &  \\
\bottomrule
\end{tabularx}
\end{sc}
\end{small}
\end{center}
\caption{Orbital similarity between different types of nonlinear systems under linear coordinate transformation. Similarity was shown in mean $\pm$ standard error of mean in the second column, and median $\pm$ inter-quarter range in the third column.}
\label{tab:nonlinear-sys}
\vskip -0.1in
\end{table}

\subsubsection{Nonlinear coordinate transformation between linear systems}

Next, we examined whether DFORM can learn a \textit{nonlinear} coordinate transform that more strongly warps the phase space. It is well-known that two non-degenerate linear dynamical systems are topologically equivalent if they have the same number of eigenvalues on the left (or right) half complex plane. However, in some cases, such as between a stable node (with only negative real eigenvalues) and a stable focus (with only complex eigenvalues in the left half plane), the orbit-matching homeomorphism is not differentiable at the origin. Since the DFORM architecture is bi-Lipschitz (thus only having bounded rate-of-change), we expect it to have more difficulty evaluating the topological equivalence between two such systems.

Therefore, we constructed topologically equivalent 32-dimensional linear systems of the form $\dot x = Ax$ that either: (1) are similar to each other by an orthogonal transformation (i.e., of the form $A_{2} = QAQ^{-1}$); (2) are similar to each other by an invertible transformation; (3) have the same number of positive/negative/zero eigenvalues and same number of complex eigenvalues with positive/negative/zero real parts; or (4) have the same number of eigenvalues on the right half plane, left half plane and imaginary axis, while not requiring the total number of real eigenvalues to be the same for the two systems. We simulated 30 pairs of systems for each type, and trained DFORM for 3000 batches. The sample distribution was chosen as the standard normal distribution to emphasize the only fixed point.

\begin{figure*}[t]
\vskip 0.2in
    \centering
    \includegraphics[width=\textwidth]{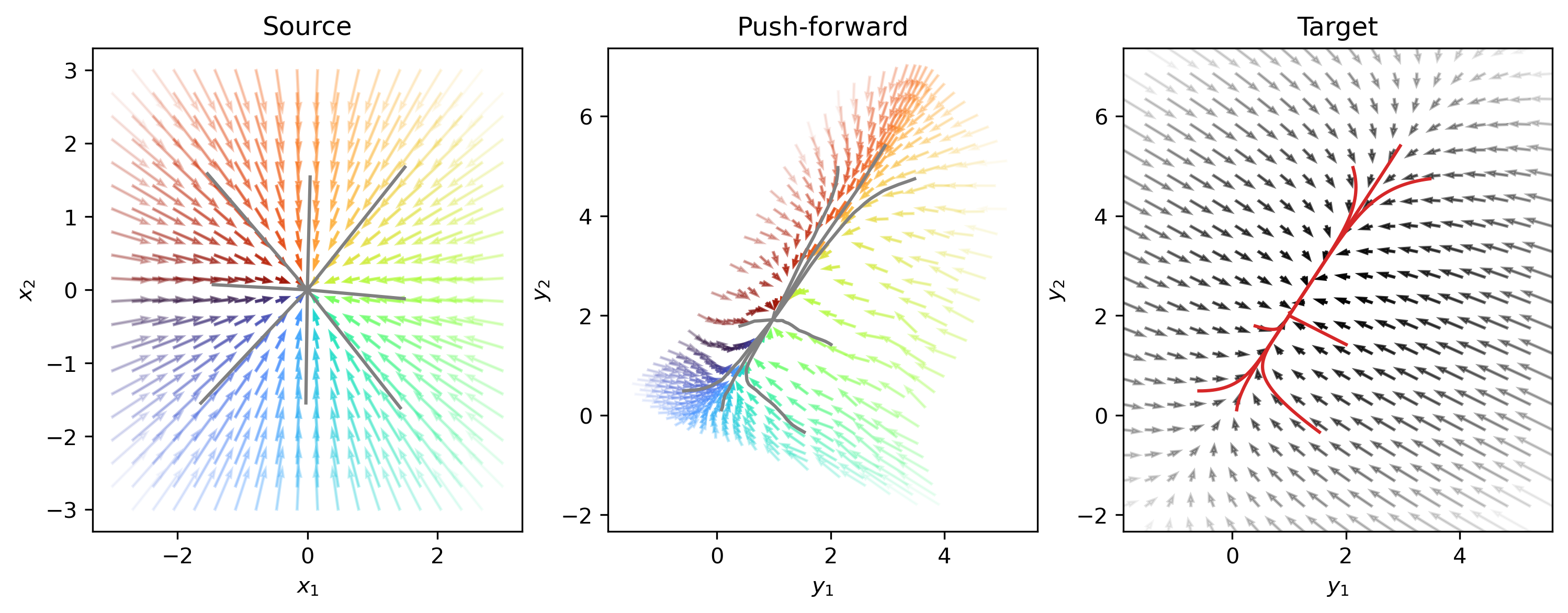}
    \caption{Alignment between two-dimensional systems. The DFORM network used here contains 40 instead of 10 layers. Left: vector field (arrows) and simulated trajectories (gray lines) for the first system. Middle: push-forward of the vector field and trajectories by the DFORM model. Right: vector field and trajectories of the second system. Initial conditions for the trajectories were the same as in the middle panel.}
    \label{fig:Linear2D}
\vskip -0.2in
\end{figure*}

Results are shown in \cref{tab:eig-equiv}. 
DFORM was still able to produce above 0.9 similarity in the most difficult case. An example for alignment between two planar systems is shown in \cref{fig:Linear2D}, where we intentionally also include a bias term in the second system. Note that DFORM not only matches the global shift and rotation, but also locally strongly `deforms' the space, producing strong alignment of the push-forward and target vector fields, thus revealing similarity of the source and target dynamics.

\begin{table}[htb]
\vskip 0.15in
\begin{center}
\begin{small}
\begin{sc}
\begin{tabularx}{.99\columnwidth}{lcccr}
\toprule
Type & Similarity \\
\midrule  
Orthogonal transformation & $0.981 \pm 0.001$ \\
Invertible transformation & $0.950 \pm 0.001$ \\
Same sign \& type of eigenvalues & $0.946 \pm 0.002$ \\
Same sign only & $0.920 \pm 0.004$ \\
\bottomrule
\end{tabularx}
\end{sc}
\end{small}
\end{center}
\caption{Orbital similarity between different types of equivalent linear systems. Similarity was shown in mean $\pm$ standard error of mean.}
\label{tab:eig-equiv}
\vskip -0.1in
\end{table}

\subsection{Orbital similarity generalizes smooth orbital equivalence}

\begin{figure}[htb]
\vskip 0.2in
\begin{center}
    \centerline{\includegraphics[width=0.7\columnwidth]{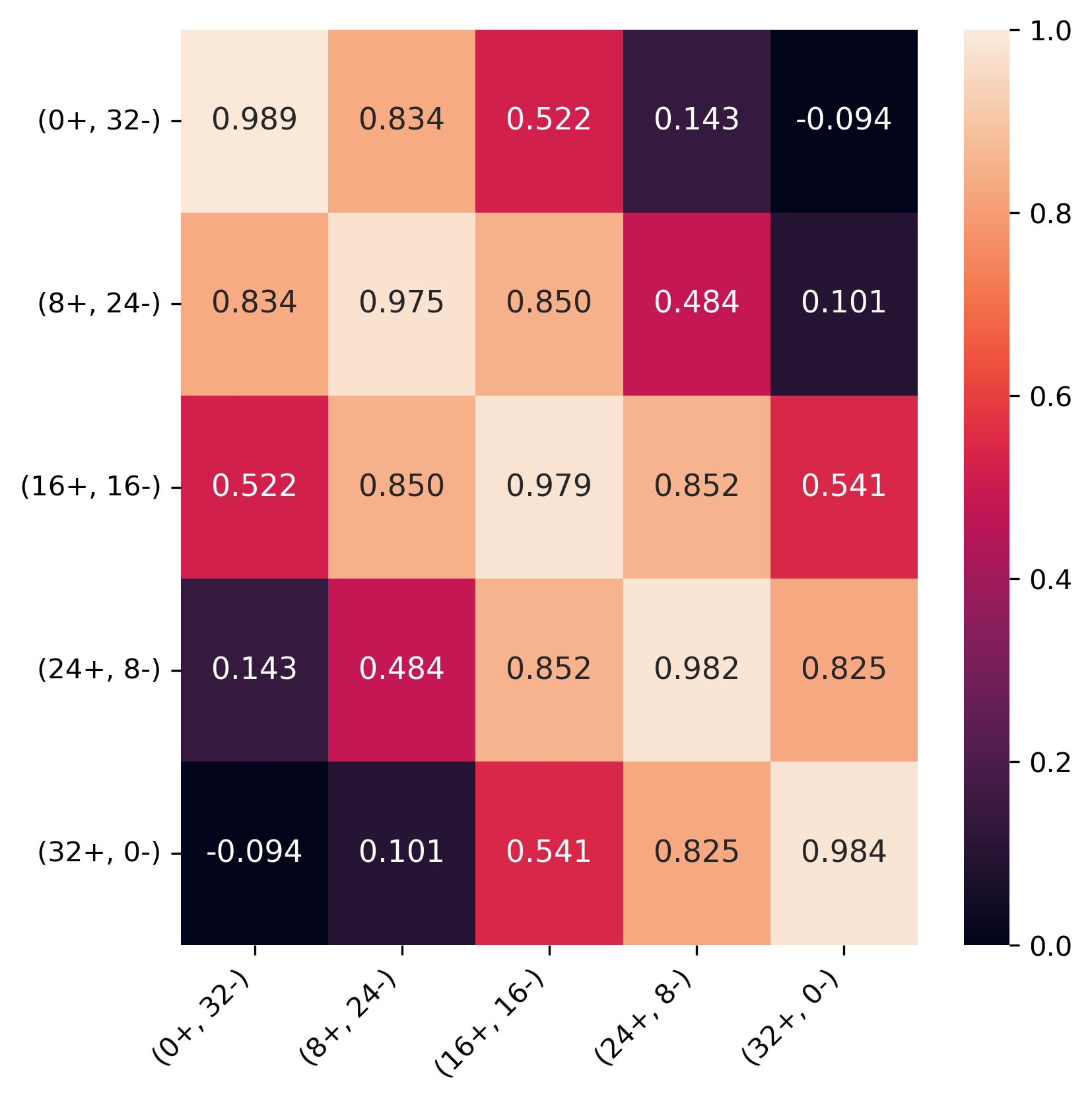}}
    \caption{Pairwise similarity between five groups of linear systems. Each group has a different number of positive and negative eigenvalues as indicated in the labels. Each block represents the average similarity over 15 different pairs. Note: the upper and lower triangle are the same.}
    \label{fig:eig-sim}
\end{center}
\vskip -0.2in
\end{figure}

As a generalization of the concept of topological equivalency, we expect our similarity measure to also depend on the signs of the real parts of the eigenvalues. To confirm this hypothesis, we generated 32-dimensional linear dynamical systems with nonzero real eigenvalues, ranging from all negative to all positive. The pairwise similarity between these systems are shown in \cref{fig:eig-sim}. As expected, the similarity between two linear systems were determined by the similarity between the signs of their eigenvalues (\cref{sgn-diff}).

\begin{table}[htb]
\caption{Orbital similarity between linear systems with different eigenvalue sign distribution. Similarity is shown as mean $\pm$ standard error of mean.}
\label{sgn-diff}
\vskip 0.15in
\begin{center}
\begin{small}
\begin{sc}
\begin{tabularx}{.99\columnwidth}{lcccr}
\toprule
Proportion of eigenvalues & Similarity \\
with same sign & \\
\midrule  
$100\%$ & $0.982 \pm 0.001$ \\
$75\%$ & $0.840 \pm 0.004$ \\
$50\%$ & $0.515 \pm 0.009$ \\
$25\%$ & $0.122 \pm 0.012$ \\
$0\%$ & $-0.094 \pm 0.008$ \\
\bottomrule
\end{tabularx}
\end{sc}
\end{small}
\end{center}
\vskip -0.1in
\end{table}

\subsection{DFORM clarifies dynamical similarity across learned RNNs}

Finally, we used an illustrative example to demonstrate how DFORM can characterize the dynamics of different networks trained on the same task, thus clarifying their generative mechanisms. 

\subsubsection{Task paradigm}
\begin{figure}[htb]
\vskip 0.2in
\begin{center}
    \centerline{\includegraphics[width=\columnwidth]{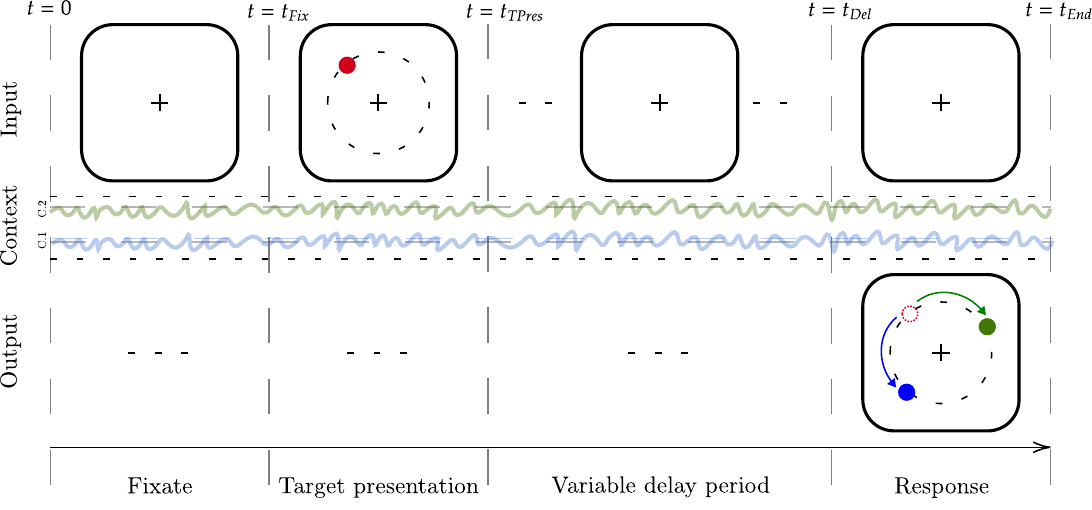}}
    \caption{Memory task for DFORM analysis. Networks maintain fixation around the origin for a duration of time denoted as $t = T_{Fix}$. Subsequently, a stimulus (highlighted in red) is introduced to the network and remains visible until the time $t = T_{TPres}$. Following this presentation, a variable delay period ensues, concluding with a predetermined response time, during which the network produces a `saccade' to the intended, context-dependent location (e.g., green vs. blue).}
    \label{fig:SWMT}
\end{center}
\vskip -0.2in
\end{figure}

We considered a typical center-out delayed match to sample spatial memory task (remember and output a location on a circle), where dynamics are thought to play a key role in representing the position of the stimulus. We enriched this example by adding a persistent contextual signal that dictates a rotation that must be applied to the output (\cref{fig:SWMT}). 
We considered two architectures for encoding this contextual information. In the first scenario, we adopted a `vanilla' structure, represented as 
\begin{equation}
\tau \dot{x} = -x + W \tanh(x) + B_1 u_s + B_1 u_c
\end{equation}
where the input vector $u_c$ provides the context signal (e.g., as a hot vector). The vector $u_s$ includes the stimulus as well as fixation signal. We refer to this as a v-type model. Conversely, in the second scenario, we used a connectivity-switched architecture, denoted as
\begin{equation}
\tau \dot{x} = -x + (W + \Gamma_i) \tanh(x) + B_1 u_s.
\end{equation}
Here, each context is delivered via a low-rank matrix $\Gamma_i$, which additively modifies the network connectivity weights (we refer to this as a w-type model). Both models are readily trained via conventional gradient approaches. We emphasize that we are not arguing in favor or against these architectures, nor are we purporting any result about their efficacy \textit{per se}. We are simply using them as structurally distinct RNN implementations whose emergent dynamics we can then compare.

\subsubsection{DFORM setup}
We use DFORM to analyze the dynamics of the learned models. We obtained 10 models each of the v- and w-types. 
Here we will focus our analysis on the response period, where the networks will be modulated by the context as well as the task fixation signal. 
For DFORM sampling, in this case we used asymptotic distribution sampling with white noise of magnitude $1.5$, based on simulations of 1000 trials.
The sample distribution surrounds the limit sets of the model and overlays regions of state space that the model traverses during actual simulation of trials \cref{fig:memory-smp}. Therefore, such a sample distribution prioritize the region-of-interest in both dynamical and functional sense.

We trained a DFORM network for 6000 batches of samples for each pair of different networks. The highest orbital similarity across three repetitions was taken as the similarity between two networks.

\subsubsection{SVCCA setup}
To compare our results with existing methods, we implemented the aforementioned singular vectors canonical correlations analysis (SVCCA) framework \cite{raghu2017svcca}. 
We performed 1000 realizations of each model, thus creating a set of trajectories, then performed PCA dimensionality reduction to retain 95\% of explained variance (as suggested in SVCCA). We then projected the reduced trajectories back to the original network phase space. We performed a canonical correlation analysis to align the trajectories of different models and obtained the correlation coefficient. The analysis was implemented using Python's \textit{scikit-learn} package.

\subsubsection{DFORM analysis indicates common functional dynamics}

\begin{figure}[htb]
\vskip 0.2in
\begin{center}
    \centerline{\includegraphics[width=\columnwidth]{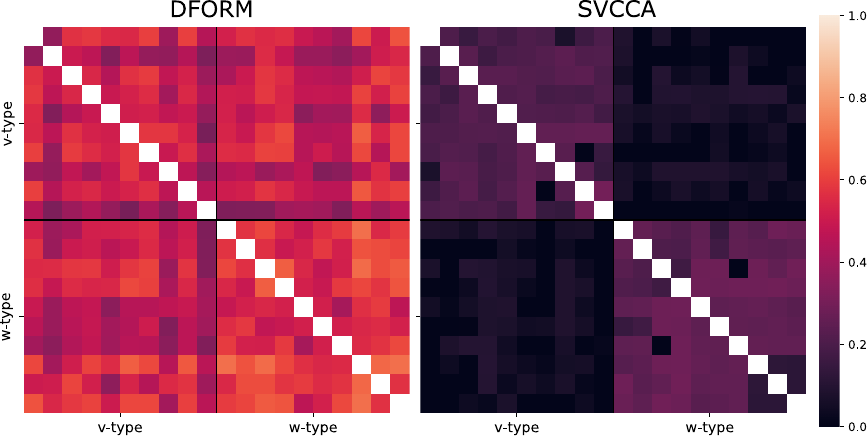}}
    \caption{Similarity between models computed by DFORM and SVCCA. Note: the upper and lower triangle are the same.}
    \label{fig:MemoryRNN}
\end{center}
\vskip -0.2in
\end{figure}

The similarity measured by DFORM and SVCCA is shown in \cref{fig:MemoryRNN}.
As expected, SVCCA draws a clear distinction between the v- and w-types of networks, since the trajectories of these models are quite distinct in state space (due to the fundamental difference in the way the cue enters the vector field). 
However, DFORM paints a different picture on the dynamics of these models, and specifically that their vector fields are in fact quite similar despite overt differences in architecture (\cref{fig:MemoryRNN}).
This is consistent with the fact that most networks we examined, regardless of architecture, seem to have developed a ring-shape slow manifold structure connecting stable fixed points (\cref{fig:memory-traj}). Such a structure is well-known to emerge in RNNs trained on this type of task \cite{khona2022attractor}.
Indeed, although different networks might have a different number of fixed points, their behavior under noise might be functionally equivalent, as suggested by the ring-shape asymptotic distributions that occurs in many networks with different number of attractors \cref{fig:memory-smp}. 
Thus, despite a few individual variations (e.g., v-type network 10), a common dynamical mechanism seems to be embedded in these networks, despite distinctions architecture and attractor landscapes.

\section{Conclusion \& discussion}
We propose a general framework, DFORM, to enable the comparison of learned dynamics. The framework is based on the dynamical systems definition of smooth orbital equivalence, and operates by attempting to learn a diffeomorphism that maps between the vector fields of two models under consideration. To implement this framework, we introduce a new loss function as well as a network architecture and learning scheme for its minimization. The outcome is a continuous similarity metric that quantifies a alike the dynamics are of any two learned dynamical models (e.g., RNNs). As such, DFORM may be quite generally useful in clarifying commonalities in how different learned models use their dynamics in the service of tasks. Beyond its use in applicative contexts, moving forward, several extensions of this framework may be of interest, including allowing for comparisons of systems of different dimensionality. Besides, explicitly regularizing of the network to match the push-forward and target \textit{sample distributions}, or alternatively, factoring the similarity between the two distributions into the orbital similarity measure, could also be interesting from both applicative and theoretical standpoints.

\section{Impact statement}

This paper presents work whose goal is to advance the field of Machine Learning. There are many potential societal consequences of our work, none which we feel must be specifically highlighted here.

\bibliography{main}
\bibliographystyle{icml2024}

\newpage
\appendix
\onecolumn

\section{More examples of planar systems}

\begin{figure*}[htb]
\vskip 0.2in
\begin{center}
    \centerline{\includegraphics[width=0.8\textwidth]{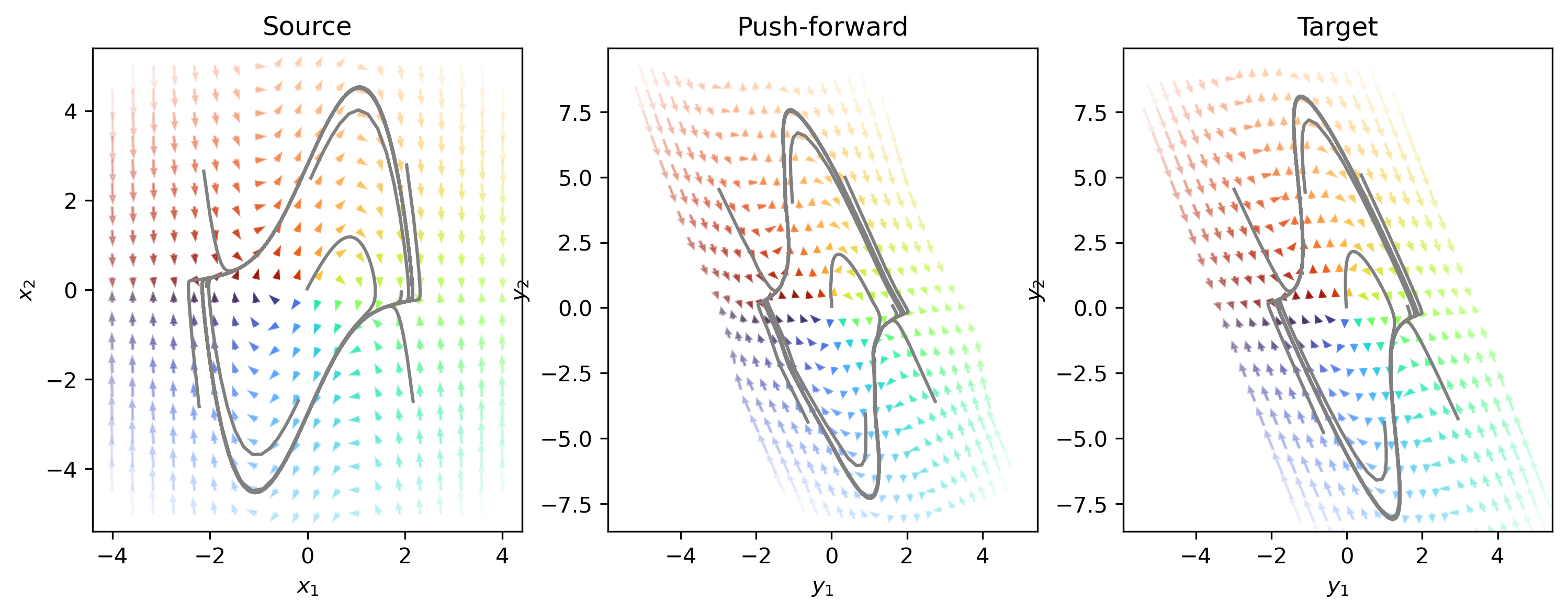}}
    \caption{Example for alignment between linearly-transformed nonlinear systems (Van der Pol oscillators). Also see \cref{fig:Linear2D}.}
    \label{fig:VDP}
\end{center}
\vskip -0.2in
\end{figure*}

\section{Simulated trajectories of memory networks}

\begin{figure}[htb]
    \centering
    \includegraphics[trim={0 0 0 1cm}, clip, width=0.95\textwidth]{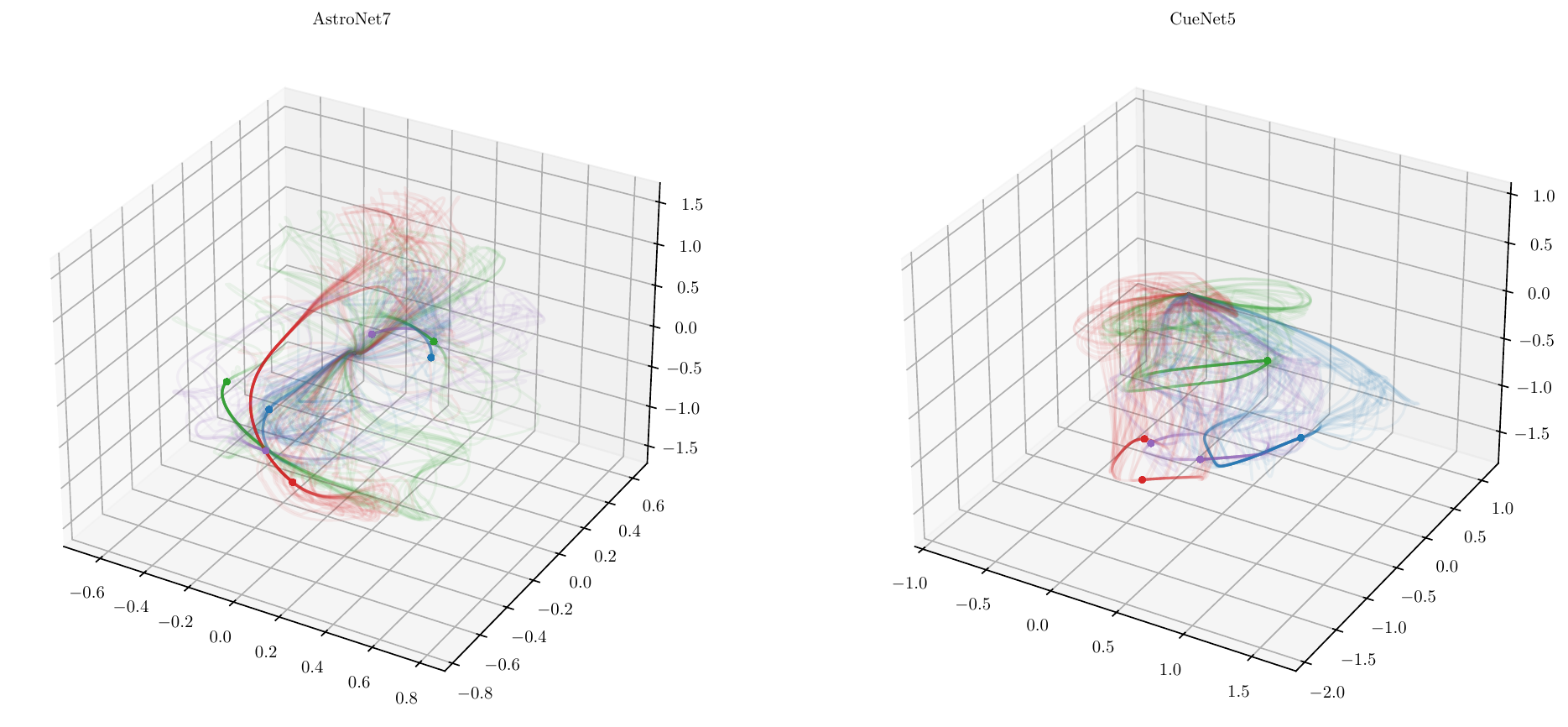}
    \caption{Simulated trajectories for two example networks. The left one is w-type and the right one is v-type. X, Y, Z axes represent the first three principal components. Contexts are indicated by different colors. Each line represents one trial with a different stimulus input. The response period was extended indefinitely until the network converged to the attractors. Note that both network showed slow manifolds around the fixed point where the trajectories were particularly dense.}
    \label{fig:memory-traj}
\end{figure}

\section{Sample distribution for memory networks}

\begin{figure}[htb]
    \centering
    \includegraphics[width=0.95\textwidth]{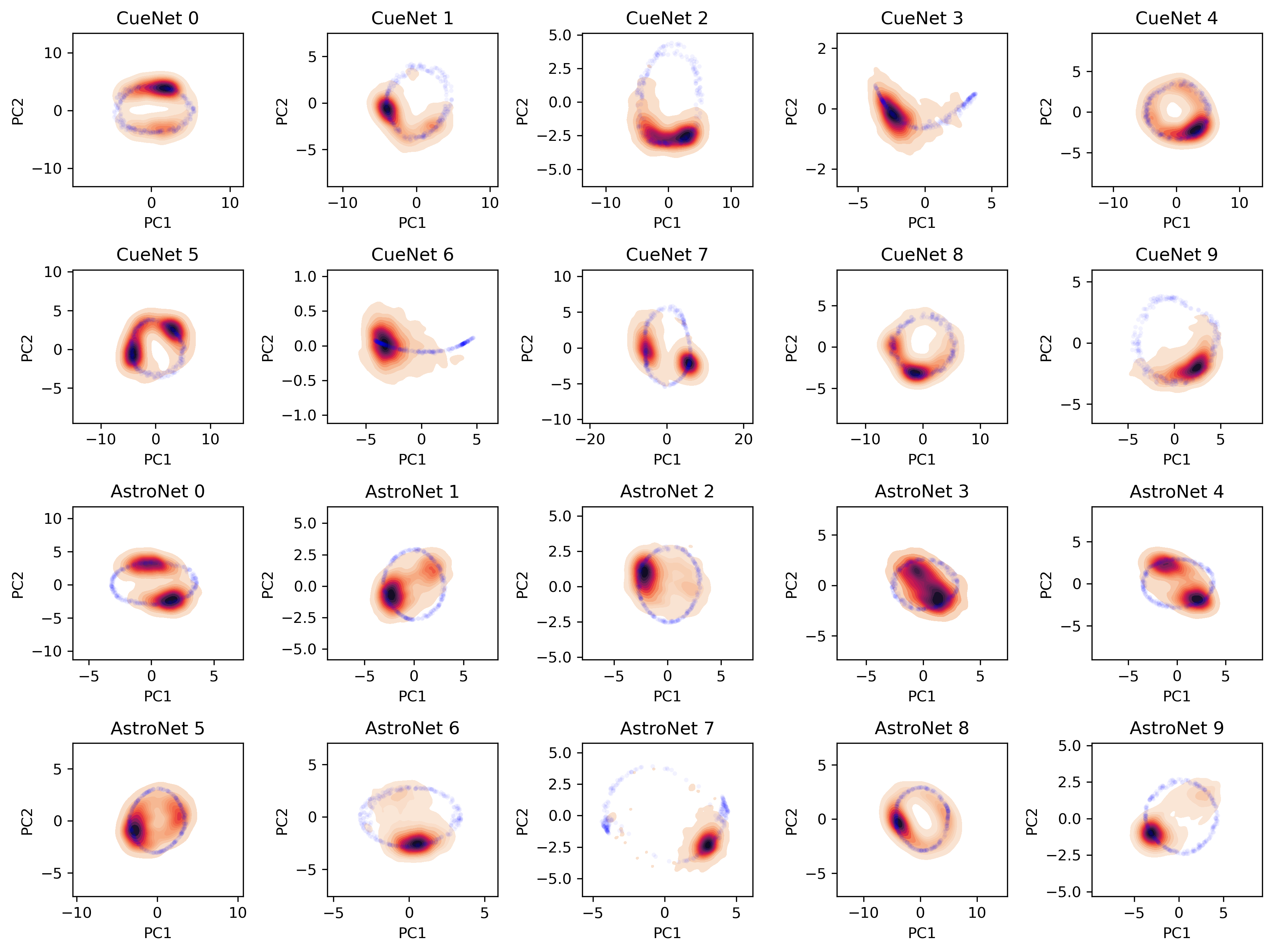}
    \caption{Sample distribution for memory networks. Each blue dot represents the final state of the network (i.e., after `saccade' in trial. PCA was applied to the set of all such final states. Density plot represents the sample distribution projected onto the first two principal components}
    \label{fig:memory-smp}
\end{figure}

\end{document}